# Neural Network Application on Foliage Plant Identification

A. Kadir
Gadjah Mada University
Yogyakarta
Indonesia

L. E. Nugroho
Gadjah Mada University
Yogyakarta
Indonesia

A. Susanto
Gadjah Mada University
Yogyakarta
Indonesia

P.I. Santosa
Gadjah Mada University
Yogyakarta
Indonesia

## ABSTRACT

Several researches in leaf identification did not include color information as features. The main reason is caused by a fact that they used green colored leaves as samples. However, for foliage plants—plants with colorful leaves, fancy patterns in their leaves, and interesting plants with unique shape—color and also texture could not be neglected. For example, *Epipremnum pinnatum 'Aureum'* and *Epipremnum pinnatum* 'Marble Queen' have similar patterns, same shape, but different colors. Combination of shape, color, texture features, and other attribute contained on the leaf is very useful in leaf identification. In this research, Polar Fourier Transform and three kinds of geometric features were used to represent shape features, color moments that consist of mean, standard deviation, skewness were used to represent color features, texture features are extracted from GLCMs, and vein features were added to improve performance of the identification system. The identification system uses Probabilistic Neural Network (PNN) as a classifier. The result shows that the system gives average accuracy of 93.0833% for 60 kinds of foliage plants.

## 1. INTRODUCTION

Several methods to identify plants based on a leaf have been introduced by several researchers. Principally, they used shape, color, or texture features, partially or completely.

Several approaches has been used to capture shape of the leaf. Tak & Hwang [1] proposed a method called Partial Dynamic Time Warping (PDTW) for comparing two curves based on Dynamic Time Warping (DTW) to matches curves. First of all, the method get distance curve from a leaf and then calculate its Fourier descriptors. Wu et al. [2] created a plant identification system that can recognize 6 kinds of plants. They used slimness ratio (ratio between major axis and minor axis), roundness ratio (ratio between area to perimeter of the leaf), solidity (expresses the degree of splitting depth in a leaf), invariant moments to represent shape of the leaf, three features to represent leaf dent, and two features to represent leaf vein. Lee & Chen [3] proposed a region-based classification method for leaves. They used slimness ratio (called aspect ratio), compactness (same as roundness ration), centroid, and horizontal and vertical projections. Then, a dissimilarity measure was used to match scores. Cheng et al. [4] used a method that combine leaf shape, leaf apex, leaf base, leaf margin, leaf phyllotaxy, leaf vein and leaf size. Pauwels et al. [5] used several shape features, including indentation depth (the distance from leaf contour to the convex hull as a function contour arc-length, i.e. the distance covered along the leaf's perimeter). Yahiaoui et al. [6] created shaped-based image retrieval in botanical collections. They proposed a descriptor called Directional Fragment Histogram (DFH). Wu et al. [7] combined geometric features and vein features to identify 32 kinds of green leaf plants. Im et al. [8] used polygonal approximation to recognize plant species of Acer family. Other method called angle code histogram (ACH) was proposed by An-Xiang et al. [9] to retrieve flower images.

Although color was neglected in several researches, several researchers have been included it to recognize plants. Zhang & Zhang [10] incorporated color features for tobacco leaf grading purpose. They used variance of red, green, and blue channel of the digital image of tobacco leaves. Man et al. [11] involved color moments that used mean, standard deviation, and skewness to identify plants.. Kebapci et al. [12] presented a content-based image retrieval system for plant image retrieval that included color features.

Texture is an important aspect in foliage plants that have beautiful pattern in their leaves. Actually, texture features were also used in several researches in plant recognition, such as in [12], that used Gabor wavelets to extract the plant texture and in [13] that tried to use Gray-Level Co-occurrence Matrix (GLCM) only to classify plants.

The plants identification systems were implemented in different ways. Several researches [2] [7] used neural networks and the others used matching strategies. General Regression Neural Network was used by Zulkifli [14], Back Propagation neural network was used by Wu et al. [2], and Probabilistic Neural Network was used by Wu et al. [7]. Man et al. [11] and Singh et al. [15] utilized Support Vector Machine (SVM) to classify plants. Cheng et al. [4] used fuzzy function to obtained scores of all characteristics, where the higher the score is, the closer to the leaf of query.

This research was the extension of our previous works [16][17]. The goal was to create an identification system that can recognize not only green leaves, but also colored leaves, that are commonly contained in foliage plants. Shape, color, texture, and also vein features are important aspects that were included in the testing. Shape features are represented by Fourier descriptors that are obtained from Polar Fourier Transform and three kinds of geometric features. PFT was used because of its good performance compared to invariant moment and Zernike moments [16]. In this experiment, 4 kinds of color moments (mean, standard deviation, skewness, and kurtosis) were inspected. GLCM were included as texture features and three vein features derived from [7] were used to improve performance of the identification system. In implementation, PNN were used as a classifier. To test the system, two kinds of datasets were used. First dataset came from [7] that contain 32 kinds of plants and the second dataset came from us, which contain 60 of foliage plants. The results show that the system gives better accuracy then the original work when using Flavia dataset. By using dataset that contains foliage plants, the system give performance 93.0833% of accuracy.

## 2. LEAF FEATURES
All kinds of features used for testing in foliage plant identification are described here.

## 2.1 Polar Fourier Transform
Polar Fourier Transform that proposed by Zhang [18] has properties that are very useful for represents shape of objects, including leaf of plants. Features extracted from PFT are invariant under the actions of translation, scaling, and rotation.

Algorithm to obtain Fourier descriptors as follow.

**Input**: I, max_rad, max_ang
// I : grayscale image
// max_rad : maximum of radial frequency
// max_ang : maximum of angular frequency
**Output**: FD (Fourier descriptor)

1. width ← width of image
2. height ← height of image
3. Obtain the centroid (x_center, y_center)
4. Obtain maximum radius on the leaf (max_rad)
5. Do PFT:
    FOR freq_radial ← 0 TO  max_rad
        FOR freq_angular ← 0 TO max_ang
            FR(freq_rad+1, freq_ang+1) ← 0
            FI(freq_rad+1, freq_ang+1) ← 0
            FOR x ← 1 TO width
                FOR 1 ← 0 TO height
                    radius ← ((x-x_center)^2 +
                            (y-y_center)^2)^.5
                    theta ← atan2((y-y_center),(x-x_center))

                    IF (theta<0)
                        theta ← theta + 2 * 3.14
                    END

                    FR(freq_rad+1, freq_ang+1) ←
                        FR(freq_rad+1, freq_ang+1) +
                        I(y, x) * COS(2 * 3.14 * freq_rad *
                            (radius/max_rad) +
                            freq_ang * theta)
                    FI(freq_rad+1, freq_ang+1) ←
                        FI(freq_rad+1, freq_ang+1) −
                        I(y, x) * SIN(2 * 3.14 * freq_rad *
                            (radius/max_rad) +
                            freq_ang * theta)
                END
            END
        END
    END

6. Compute Fourier descriptors (FD):

    FOR freq_rad ← 0 TO max_rad
        FOR freq_ang ← 0 TO max_ang
            IF (rad==0) && (ang==0)
                dc ← (FR(1,1)^2 + FI(1,1)^2)^.5
                FD(1) ← dc/(3.14 * rad_maks^2);
            ELSE
                FD(freq_rad * n+freq_ang+1) ←
                    (FR(freq_rad+1,freq_ang+1)^2 +
                    FI(freq_rad+1, freq_ang+1)^2)^.5 / dc;
            END
        END
    END

The centroid (x_c, y_c) calculated by using formula:

$$x_c = \frac{1}{M}\sum_{x=0}^{M-1} x, y_c = \frac{1}{N}\sum_{x=0}^{N-1} y, \qquad (1)$$

In this case, (r, T ) is computed by using:

$$r = \sqrt{(x-x_c)^2 + (y-y_c)^2}, \theta = \arctan\frac{y-y_c}{x-x_c} \qquad (2)$$

The maximum radius on the leaf is obtained by generating contour of the leaf first. For example, the Canny edge detection operator can be used for that purpose. Then, the radius is calculated by measure the distance between a pixel in the contour  and the centroid. Brute force method [19] can be used to search for the maximum distance between every pair of points that constitute the leaf. In our experiment, maximum of radial frequency = 4 and maximum of angular frequency = 6.

## 2.2 Geometric Features
There are three kinds of geometric features involved as shapes features: slimness ratio, roundness ratio, and dispersion. Slimness ratio (sometime called as aspect ratio or eccentricity) is the ratio of the length of minor axis to the length of major axis. Roundness ratio (or circularity ratio) is the ratio of the area of a shape to the shape's perimeter square. Dispersion is ratio between the radius of the maximum circle enclosing the region and the minimum circle that can be contained in the region [20]. The three features can be described mathematically as follows.

$$eccentricity = \frac{w}{l} \qquad (3)$$

$$roundness = \frac{A}{P^2} \qquad (4)$$

$$dispersion = \frac{\max(\sqrt{(x_i - \bar{x})^2 + (y_i - \bar{y})^2})}{\min(\sqrt{(x_i - \bar{x})^2 + (y_i - \bar{y})^2})} \qquad (5)$$

where w is the length of the leaf's minor axis, l is the length of the leaf's major axis, A is the area of leaf, P is perimeter of the leaf, $(\bar{x}, \bar{y})$ is the centroid of the leaf, and (x_i, y_i) is the coordinate of a pixel in the leaf contour.

## 2.3  Color Features

Color features on a leaf can be extracted by using statistical calculations such as mean, standard deviation, skewness, and kurtosis [21]. Those calculations are applied to each component in RGB color space.

Mean or expected value provides a measure of distribution. It is calculates as follow.

$$\mu = \frac{1}{MN} \sum_{i=1}^{M} \sum_{j=1}^{N} P_{ij} \qquad (6)$$

Variance is a measure of dispersion in the distribution, The square root of the variance is called standard deviation. Formula to obtain standard deviation is shown in Eq. 7.

$$\sigma = \sqrt{\frac{1}{MN} \sum_{i=1}^{M} \sum_{j=1}^{N} (P_{ij} - \mu)^2} \qquad (7)$$

Skewness is a measure of symmetry. Distributions that are skewed to the left will have negative coefficient of skewness, and distributions that are skewed to the right will have a positive value. If the distribution is symmetric then the coefficient of skewness is zero. The Skewness is computed as below:

$$\theta = \frac{\sum_{i=1}^{M} \sum_{j=1}^{N} (P_{ij} - \mu)^3}{MN\sigma^3} \qquad (8)$$

Kurtosis is a measure of whether the data are peaked or flat relative to normal distribution. It is computed by using following formula:

$$\gamma = \frac{\sum_{i=1}^{M} \sum_{j=1}^{N} (P_{ij} - \mu)^4}{MN\sigma^4} - 3 \qquad (9)$$

On Eq. 6 through Eq. 9, M is the height of the image, N is the width of the image, and $P_{ij}$ is the value of color on rows i and column j.

## 2.4  Texture Features

Texture features are extracted from GLCMs. GLCM is very useful to obtain valuable information about the relative position of the neighbouring pixels in an image [23]. The co-occurrence matrix GLCM(i,j) counts the co-ocurrence of pixels with grey value i and j at given distance d. The direction of neighboring pixels to represents the distance can be selected, for example 135°, 90°, 45°, or 0°, as illustrated in Fig. 1. Fig. 2 illustrates how to generate the matrix using direction 0° and distance between pixels is equal 1.

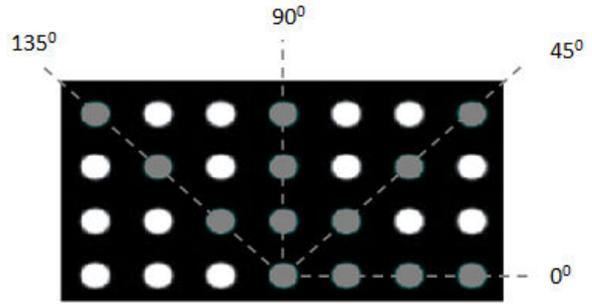

**Fig 1: Direction in calculating GLCM**

By adding its transpose, a simmetric matrix will be obtained, as shown in Fig. 3. However, the result is still unnormalized. Therefore, the normalization process should be done in order to remove dependence on the image size, by arranging all elements in the matrix so the total of all element values equals 1. The result is shown in Fig. 4.

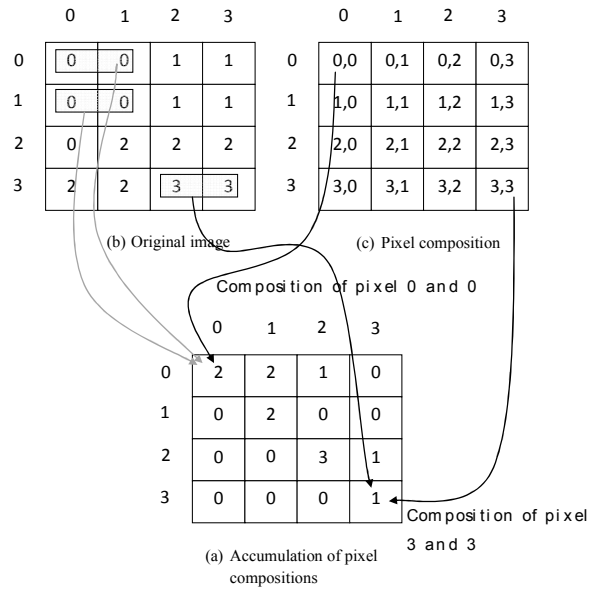

**Fig 2: First step to generate GLCM**

$$\begin{bmatrix} 2 & 2 & 1 & 0 \\ 0 & 2 & 0 & 0 \\ 0 & 0 & 3 & 1 \\ 0 & 0 & 0 & 1 \end{bmatrix} + \begin{bmatrix} 2 & 0 & 0 & 0 \\ 2 & 2 & 0 & 0 \\ 1 & 0 & 3 & 0 \\ 0 & 0 & 1 & 1 \end{bmatrix} = \begin{bmatrix} 4 & 2 & 1 & 0 \\ 2 & 4 & 0 & 0 \\ 1 & 0 & 6 & 1 \\ 0 & 0 & 1 & 2 \end{bmatrix}$$

Transpose       GLCM before normalization

**Fig 3: Procedure to create symmetric matrix**

$$\begin{bmatrix} \dfrac{4}{24} & \dfrac{2}{24} & \dfrac{1}{24} & \dfrac{0}{24} \\ \dfrac{2}{24} & \dfrac{4}{24} & \dfrac{0}{24} & \dfrac{0}{24} \\ \dfrac{1}{24} & \dfrac{0}{24} & \dfrac{6}{24} & \dfrac{1}{24} \\ \dfrac{0}{24} & \dfrac{0}{24} & \dfrac{1}{24} & \dfrac{2}{24} \end{bmatrix}$$

**Fig 4: Normalized matrix**

GLCMs are usually computed for a number of different offsets unless a priori information is available about the underlying texture [22]. A common choice is to compute GLCMs for a distance of one (i.e., adjacency) and four directions, 0, 45, 90, and 135 degrees. Then, texture features are extracted from those matrices by using several scalar quantities proposed by Haralick. Newsam & Kamath [22] proposed five GLCMs derived features: angular second moment (ASM), contrast, inverse different moment (IDM), entropy, and correlation.

The ASM (or energy) is computed as

$$ASM = \sum_{i=1}^{L} \sum_{j=1}^{L} \left( GLCM(i,j)^2 \right) \qquad (10)$$

The ASM measures textural uniformity (i.e. pixel pairs repetition) [23]. The ASM has the highest value when the distribution of the grey levels constant or periodic.

The contrast is computed as

$$Contrast = \sum_{i=1}^{L} \sum_{j=1}^{L} (i-j)^2 \left( GLCM(i,j) \right) \qquad (11)$$

The contrast measures the coarse texture or variance of the grey level. The contrast is expected to be high in coarse texture, if the grey level of contrast corresponds to large local variation of the grey level [24].

The IDM (or homogeneity) is computed as

$$IDM = \sum_{i=1}^{M} \sum_{j=1}^{N} \frac{\left( GLCM(i,j) \right)^2}{1 + (i-j)^2} \qquad (12)$$

The IDM measures the local homogeneity a pixel pair. The homogeneity is expected to large if the grey levels of each pixel pair are similar [24].

The entropy is computed as

$$Entropy = -\sum_{i=1}^{L} \sum_{j=1}^{L} GLCM(i,j) \, x \log \left( GLCM(i,j) \right) \qquad (13)$$

The entropy measures the degree of disorder or non-homogeneity of image. Large values of entropy correspond to uniform GLCM. For texturally uniform image, the entropy is small.

The correlation texture measures the linear dependency of gray levels on those of neighboring pixels. This feature is computed as

$$Correlation = \sum_{i=1}^{L} \sum_{j=1}^{L} \frac{(ij)(GLCM(i,j) - \mu_1'\mu_2')}{\sigma_i'\sigma_j j'} \qquad (14)$$

where

$$\mu_i' = \sum_{i=1}^{L} \sum_{j=1}^{L} i * GLCM(i,j) \qquad (15)$$

$$\mu_j' = \sum_{i=1}^{L} \sum_{j=1}^{L} j * GLCM(i,j) \qquad (16)$$

$$\sigma_i^2 = \sum_{i=1}^{L} \sum_{j=1}^{L} GLCM(i,j)(i - \mu_i')^2 \qquad (17)$$

$$\sigma_j^2 = \sum_{i=1}^{L} \sum_{j=1}^{L} GLCM(i,j)(i - \mu_i')^2 \qquad (18)$$

In Eq. 10-18, L is number of rows/columns in GLCM.

The weakness of GLCM is its dependency to rotation. Therefore, to achieve rotational invariant features, the GLCM features corresponding to four directions (135°, 90°, 45°, or 0°) are firstly calculated and then average them [22]. In our implementation, we used eight directions (135°, 90°, 45°, or 0°, 225°, 270°, 180°, or 315°) and then average them.

## 2.5 Vein Features

Vein features are obtained by using morphological opening [7]. That operation is performed on the gray scale image with flat, disk-shaped structuring element of radius 1, 2, 3, 4 and subtracted remained image by the margin. As a result, a structure like vein is obtained. Fig. 5 shows an example of vein resulted by such operation.

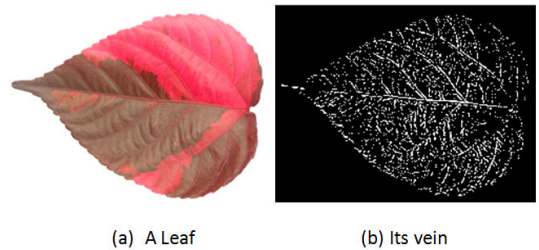

(a)  A Leaf          (b) Its vein

**Fig 5: Illustration of vein processed by using morphological operation**

Based on that vein, 4 features are calculated as follow:

$$V_1 = A_1 / A, V_2 = A2 / A, V_3 = A_3 / A, V_4 = A_4 / A \qquad (19)$$

In this case, $V_1$, $V_2$, and $V_3$ represent features of the vein, $A_1$, $A_2$, and $A_3$ are total pixels of the vein, and $A$ denotes total pixels on the part of the leaf.

## 3. IDENTIFICATION SYSTEM & PNN

Flow diagram of proposed scheme is shown in Fig. 6. After capturing the leaf image, the leaf is separated from its background by using segmentation in gray scale image. Then, all features needed are extracted. For training the neural network, all features of training samples are inputted to the PNN. The next step, weights of the neural network that are obtained from training phase will be used to test the network. The results are compared to original classes to get performance of the system.

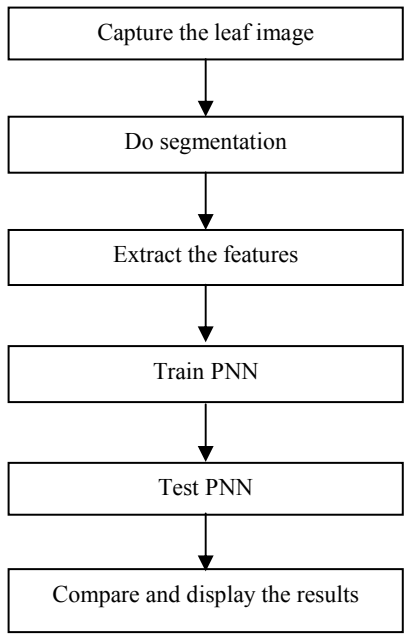

Fig 6: Samples of leaves

PNN is an implementation of statistical algorithm called kernel discriminant analysis, in which the operations are organized into a multilayered feed-forward network with four layers: 1) input layer, 2) pattern layer, 3) summation layer, and 4) output layer [25]. When an input is presented, the first layer computes distances from the input vector to the training input vectors and produces a vector whose elements indicate how close the input is to a training input. The second layer sums these contributions for each class of inputs to produce as its net output a vector of probabilities [26]. Finally, a compete transfer function on the output of the second layer picks the maximum of these probabilities, and produces a 1 for that class and a 0 for the other classes. The architecture for this system is shown in Fig. 7.

PNN has advantages: 1) fast training process, 2) an inherently parallel structure, 3) guaranteed to converge to an optimal classifier as the size of the representative training set increases, and 4) training samples can be added or removed without extensive retraining. However, PNN also has disadvantages: 1) not as general as back-propagation, 3) large memory requirement, 3) slow execution of the network and 4) requires a representative training set [25].

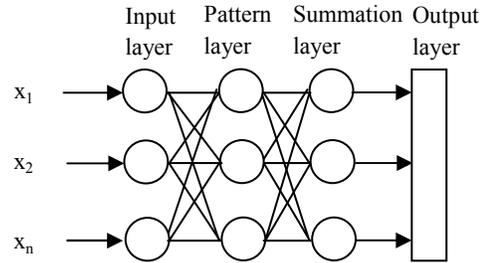

Fig 7: Architecture of PNN

According to Emary & Ramakrishnan [27], the training set of PNN must be done through representative of the actual population for effective classification. When the training size increases, PNN asymptotically converges to the Bayes optimal classifier. Mathematically, the probability density function is expressed as:

$$p(x \mid w_j) = \frac{1}{(2\pi)^{d/2} \sigma^d n_j} \sum_{k=1}^{n_j} \exp\left(-\frac{(x - X_k)^2}{2\sigma^2}\right) \quad (20)$$

where $p(x \mid w_j)$ represents the conditional probability x to class $w_j$, $x$ is input vector, $X_k$ is training dataset, $d$ is the number of the input vector, $n_j$ is the number of samples for class j, $\sigma$ is smoothing factor that its value is inputted heuristically [25].

Fig. 8 shows that correct classification becomes optimal only in certain smoothing factor. In the experiment, $\sigma = 0.05$ was chosen.

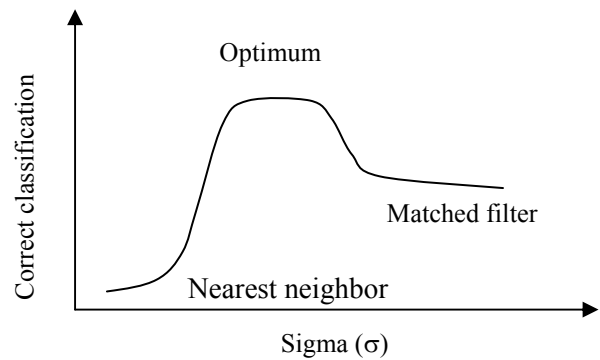

Fig 8: Relation between correct classification and smoothing factor

In order to measure performance of the system, the following formula was used:

$$Performance = \frac{n_r}{n_t} \qquad (21)$$

where $n_r$ is relevant number of images and $n_t$ is the total number of query.

## 4. EXPERIMENTAL RESULTS

To test the system, two kinds of dataset had been used. Flavia dataset was used to compare the proposed method to the original work [7]. The second dataset, called Foliage dataset, was prepared to test leaves with various colors and patterns. Flavia dataset contains 32 kinds of leaves, whereas Foliage dataset contains 60 kinds of leaves. Several samples of leaves in Flavia dataset are shown in Fig. 9.

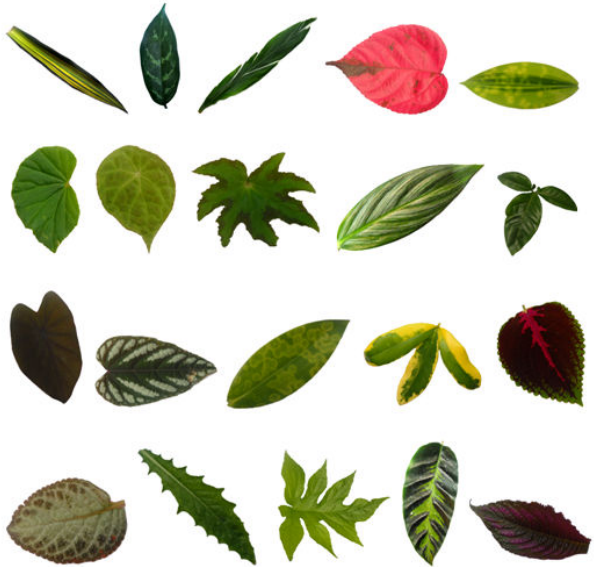

**Fig 9: Samples of leaves**

Forty leaves per species in Flavia dataset were used to train the system and ten leaves per species were used to test the system. The result is shown in Table 1. As shown, the performance of our method is slightly better than the original work.

**Table 1. Comparison of the identification system's performance using Flavia dataset**

| Method | Average Accuracy |
|---|---|
| Wu's result [7] | 90.3120% |
| Our result | 94.6875% |

When testing was done to Foliage dataset, the best result was achieved at 93.0833% of accuracy. The detail of experiment's results is shown in Table 2. In that table, number training data is equal 95 for Foliage dataset and 40 for Flavia dataset.

**Table 2. Results for Flavia and Foliage dataset**

| Features Included | Average Accuracy of Flavia | Average Accuracy of Foliage |
|---|---|---|
| PFT | 76.2500% | 72.0000% |
| PFT + 3 geometric features | 78.7500% | 75.8833% |
| PFT + 3 geometric features + 3 mean of colors | 83.4375% | 85.5833% |
| PFT + 3 geometric features + 3 mean of colors + 3 standard deviation of color | 88.4375% | 90.0833% |
| PFT + 3 geometric features + 3 mean of colors + 3 standard deviation of colors + 3 skewness of colors | 87.8125% | 91.0000% |
| PFT + 3 geometric features + 3 mean of colors + 3 standard deviation of colors + 3 skewness of colors + 3 kurtosis of colors | 88.1250% | 90.9167% |
| PFT + 3 geometric features + 3 mean of colors + 3 standard deviation of colors + 3 skewness of colors + 3 kurtosis of colors + 5 GLCM | 90.9375% | 92.3333% |
| PFT + 3 geometric features + 3 mean of colors + 3 standard deviation of colors + 3 skewness of colors + 5 GLCM | 90.6250% | 92.5000% |
| PFT + 3 geometric features + 3 mean of colors + 3 standard deviation of colors + 3 skewness of colors + 1 vein features | 92.1875% | 93.0833% |
| PFT + 3 geometric features + 3 mean of colors + 3 standard deviation of colors + 3 skewness of colors + 2 vein features | 93.4375% | 92.5833% |
| PFT + 3 geometric features + 3 mean of colors + 3 standard deviation of colors + 3 skewness of colors + 3 vein features | 94.6875% | 92.1667% |
| PFT + 3 geometric features + 3 mean of colors + 3 standard deviation of colors + 3 skewness of colors + 4 vein features | 94.0625% | 92.0833% |

Based on the results, kurtosis of colors can improve the identification system performance of Flavia data, but not for Foliage data. The same effect occurs in features vein 2, vein 3, and vein 4. Therefore, decision in choosing kurtosis, vein 2, vein 3, and vein 4 depends on the kinds of plants to be identified. For foliage plants, it seems we can exclude kurtosis.

Fig. 10 shows relation between number of samples for training in Foliage dataset and the system's performance. In this case, Fourier descriptors from PFT, 3 geometric features, 3 mean of colors, 3 standard deviation of colors, 3 skewness of colors features and 1 vein feature was used. The figure shows that the identification system's performance increases when training data increases.

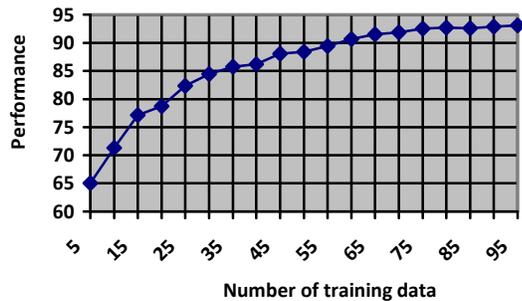

**Fig 10: Relation between number of training data and system's performance**

## 5. CONCLUSION

This paper reports the experiments in testing performance of the foliage plant identification system, which incorporated shape, vein, color, and texture features and used PNN as a classifier. The result shows that the combination of the features can improve the performance compared to the original work when testing was done by using Flavia dataset. The average accuracy is 94.6875%. The combination of the features also gives a prospective result when they applied to test foliage plants. In that case, the average accuracy was 93.0833%. However, some other works are still needed to improve the performance and it is a challenge for future researches.